\documentclass[letterpaper, 10 pt, conference]{ieeeconf}  

\IEEEoverridecommandlockouts                              

\usepackage{times}  
\usepackage{helvet}  
\usepackage{courier}  
\usepackage[hyphens]{url}  
\usepackage{graphicx} 
\usepackage[utf8]{inputenc} 
\usepackage[T1]{fontenc}    
\usepackage{url}            
\usepackage{booktabs}       
\usepackage{amsfonts}       
\usepackage{nicefrac}       
\usepackage{microtype}      
\usepackage{xcolor}         
\usepackage{subcaption}
\usepackage{amsmath} 
\usepackage{booktabs}
\usepackage{afterpage}
\usepackage{fancyhdr}
\usepackage{lipsum}
\usepackage{textcomp} 
\usepackage{marvosym}
\usepackage{caption} 
\usepackage{algorithm}
\usepackage{algorithmic}

\usepackage{newfloat}
\usepackage{listings}

\title{\LARGE \bf
DMS-Net:Dual-Modal Multi-Scale Siamese Network for Binocular Fundus Image Classification
}

\author{Guohao Huo$^{1}$\textsuperscript{$\dagger$}, Zibo Lin$^{1}$\textsuperscript{$\dagger$}, Zitong Wang$^{1}$\textsuperscript{$\dagger$}, Ruiting Dai$^{1}$, Hao Tang$^{2}$\textsuperscript{,\Letter}%
\thanks{\textsuperscript{\Letter}Corresponding author}%
\thanks{\textsuperscript{$\dagger$}These authors contributed equally.}%
\thanks{$^{1}$Guohao Huo, Zibo Lin, Zitong Wang, Ruiting Dai are with the University of Electronic Science and Technology of China. 
        {\tt\small \{gh.huo513, rtdai\}@gmail.com}}%
\thanks{$^{2}$Hao Tang is with School of Computer Science, Peking University. 
        {\tt\small hao.tang@vision.ee.ethz.ch}}%
}

\begin{document}

\maketitle
\thispagestyle{empty}
\pagestyle{empty}

\begin{abstract}
Ophthalmic diseases pose a significant global health burden. However, traditional diagnostic methods and existing monocular image-based deep learning approaches often overlook the pathological correlations between the two eyes. In practical medical robotic diagnostic scenarios, paired retinal images (binocular fundus images) are frequently required as diagnostic evidence. To address this, we propose DMS-Net—a dual-modal multi-scale siamese network for binocular retinal image classification. The framework employs a weight-sharing siamese ResNet-152 architecture to concurrently extract deep semantic features from bilateral fundus images. To tackle challenges like indistinct lesion boundaries and diffuse pathological distributions, we introduce the OmniPool Spatial Integrator Module (OSIM), which achieves multi-resolution feature aggregation through multi-scale adaptive pooling and spatial attention mechanisms. Furthermore, the Calibrated Analogous Semantic Fusion Module (CASFM) leverages spatial-semantic recalibration and bidirectional attention mechanisms to enhance cross-modal interaction, aggregating modality-agnostic representations of fundus structures. To fully exploit the differential semantic information of lesions present in bilateral fundus features, we introduce the Cross-Modal Contrastive Alignment Module (CCAM). Additionally, to enhance the aggregation of lesion-correlated semantic information, we introduce the Cross-Modal Integrative Alignment Module (CIAM). Evaluation on the ODIR-5K dataset demonstrates that DMS-Net achieves state-of-the-art performance with an accuracy of 82.9\%, recall of 84.5\%, and a Cohen's kappa coefficient of 83.2\%, showcasing robust capacity in detecting symmetrical pathologies and improving clinical decision-making for ocular diseases. Code and the processed dataset will be released subsequently.
\end{abstract}

\section{Introduction}

Ophthalmic diseases have emerged as the third leading global cause of vision loss, following cardiovascular diseases and cancer. According to statistics from the World Health Organization (WHO) \cite{WHO2019}, approximately 2.2 billion people worldwide are affected by various eye conditions. Driven by population aging and the surge in metabolic disorders \cite{riskUp}, the prevalence of common ocular diseases such as diabetic retinopathy, glaucoma, and cataracts continues to rise. However, the uneven distribution of medical resources and the inefficiency of traditional manual diagnostics heavily reliant on clinical expertise, prevent many patients from accessing timely professional diagnosis and treatment.
To address this challenge, computer vision-based ophthalmic medical robots, particularly disease diagnostic robots, are demonstrating significant potential. These systems can rapidly and automatically analyze patients' ocular images (e.g., fundus photographs, Optical Coherence Tomography (OCT) scans), assisting clinicians in identifying signs of pathology. This technology holds promise for enhancing diagnostic efficiency and accessibility, thereby alleviating shortages in healthcare resources \cite{robot1}.

Recent breakthroughs in computer vision technology have revolutionized ophthalmic diagnostics through deep learning-based image classification methods. Convolutional Neural Networks (CNNs), such as ResNet and EfficientNet, can automatically extract pathological features—like microaneurysms and hemorrhagic lesions—from fundus images via end-to-end learning \cite{CNN_Dev}. The convolution-free Vision Transformer (ViT) architecture \cite{ViT_Dev} leverages self-attention mechanisms to further enhance the recognition accuracy of complex pathologies, such as macular degeneration and choroidal neovascularization. Crucially, despite significant progress in monocular fundus image classification (e.g., the EfficientNet V4 model achieving 98.7\% accuracy in OCT image classification \cite{Eff_v4}), clinical evidence reveals that patients with unilateral ocular pathology are at substantially increased risk of developing pathological changes in the contralateral eye within subsequent years, due to shared risk factors (including genetic predisposition and oxidative stress) \cite{Ano_Eye}. Consequently, fully exploiting the underlying correlative and differential patterns in lesion spatial distribution, morphological evolution, and pathological severity between both eyes is critical for enhancing models' capabilities in precise differential diagnosis and staging of specific diseases.

To address the aforementioned challenges, we propose DMS-Net, a framework for binocular retinal image (binocular fundus image) classification. DMS-Net integrates a siamese network architecture with fundus image analysis techniques. It employs a weight-sharing Siamese ResNet-152 backbone to concurrently extract comprehensive deep semantic features from dual-channel (left and right eye) ocular images. Simultaneously, to tackle technical challenges in fundus pathological imaging, such as indistinct lesion boundaries, diffuse pathological distributions, and significant morphological variations, we developed the OmniPool Spatial Integrator Module (OSIM). OSIM achieves multi-resolution feature fusion through multi-scale adaptive pooling and spatial attention mechanisms.

Furthermore, we designed the Calibrated Analogous Semantic Fusion Module (CASFM) to enhance cross-modal (inter-channel) spatial-semantic interaction. CASFM utilizes spatial attention and an interactive attention-based feature recalibration mechanism to bridge semantic gaps between modalities and aggregate modality-agnostic representations of fundus structures.

Crucially, pathological manifestations in the two eyes often exhibit asymmetry (e.g., prominent hemorrhagic or exudative lesion features in one eye, with minimal or absent manifestations in the contralateral eye). This inherent disparity itself is vital for disease diagnosis. Thus, we innovatively introduce the Cross-modal Contrastive Alignment Module (CCAM) to identify and effectively leverage the differential semantic characteristics of lesions present in the left and right fundus images. Concurrently, considering the potential existence of correlative information for symmetrical pathologies between different lesions across both eyes, we propose the Cross-modal Integrative Alignment Module (CIAM) to strengthen the deep aggregation capability of lesion-correlated information extracted from the dual channels.

In summary, our contributions are as follows:
(1)
    We proposed DMS-Net, a novel framework specifically designed for binocular retinal image classification, employing a weight-sharing Siamese ResNet-152 backbone to extract deep semantic features from both eyes simultaneously.
(2)
    Developed the OmniPool Spatial Integrator Module (OSIM), which achieves effective multi-resolution feature fusion through multi-scale adaptive pooling and spatial attention mechanisms.
(3)
    Designed the Calibrated Analogous Semantic Fusion Module (CASFM), which employs spatial attention and an interactive attention-based feature recalibration mechanism to bridge modality-specific semantic gaps and aggregate modality-agnostic representations of fundus structures.
(4)
    Innovatively introduced the Cross-modal Contrastive Alignment Module (CCAM) to identify and leverage differential lesion semantics between left and right eyes. And proposed the Cross-modal Integrative Alignment Module (CIAM) to enhance the aggregation capability for lesion-correlated information across both eyes. 
(5)
     Demonstrated superior performance through extensive experiments on the ODIR-5K dataset, achieving an accuracy of 82.9\%, a recall of 84.5\%, and Cohen's kappa coefficient of 83.2\% for binocular fundus image classification.

\section{Related Work}

\subsection{Development of Fundus Image Classification Tasks}
Recent work on computer-aided diagnosis for fundus images ranges from classical machine-learning pipelines to deep-learning architectures. Early methods combined handcrafted preprocessing with conventional classifiers—for example, an SVM with mathematical morphology and digital-image techniques achieved high accuracy for macular-lesion detection \cite{SVM}, and hand-crafted or CNN-extracted features fed into SVM, Random Forest, MLP and J48 classifiers also obtained strong results with modest parameter counts \cite{S2020102115}. However, as feature complexity and dataset scale grow, these approaches exhibit limited generalization; similarly, simple feature-extraction schemes based on maximum principal curvature and Hessian eigenvalues fail to capture multi-scale, cross-modal information \cite{DAS2021102600}.

With deep learning, CNNs have become dominant for fundus-image classification. EfficientNet-based multi-class models perform well on small datasets but may overemphasize local lesion cues at the expense of global context \cite{CNN_Eff}; analogous shortcomings are observed for SqueezeNet-style architectures despite their multi-scale designs \cite{THANKI2023100140}. Hybrid frameworks that combine CNN locality with Transformer global attention (e.g., TransEye) mitigate some issues \cite{TransEye} and earlier MIL-VT work demonstrates the feasibility of Transformer pretraining for global modeling \cite{10.1007/978-3-030-87237-3_5}, yet key challenges such as accurate delineation of lesion boundaries remain.

Clinically, unilateral pathology substantially increases contralateral risk \cite{Ano_Eye}, but most existing methods focus on monocular classification; even studies that perform feature fusion via bottleneck modules \cite{S2020102115} lack comprehensive binocular collaborative analysis and progression prediction. Overall, prior work has advanced feature extraction, classifier integration, and architecture fusion, but limitations in cross-region feature modeling and effective multi-modal integration persist—identifying clear directions for further research.

\subsection{Siamese Network-Based Classification of Binocular Fundus Images}
Since its introduction for handwritten-signature verification \cite{Siamese}, the Siamese network architecture has proven effective for image matching and classification due to its weight-sharing design and sensitivity to similarity changes. In few-shot and fine-grained tasks, Siamese models improved plant-leaf classification by constructing a metric space with a dual-channel convolutional backbone \cite{Plant_Siamese}. In medical imaging, comparisons between single-image and Siamese inputs demonstrate superior performance and generalization of the latter for glaucoma recognition, with attention mechanisms identified as a promising optimization direction \cite{LIN2022100209}.

For binocular fundus analysis, several studies validate the Siamese paradigm for capturing interocular relationships. Direct comparisons show improved binocular collaborative feature learning over monocular models \cite{Compare_Sin_Dou}; bilinear pooling has been used successfully to capture nonlinear correlations for diabetic retinopathy grading \cite{SuggerSiamese}. Conversely, monocular detectors such as YOLOv3 are reported to neglect binocular lesion associations and to struggle with blurred boundaries and dispersed features \cite{Akella2024}.

Architectural and feature-level refinements have been proposed to address fundus-image challenges. Multimodal preprocessing (edge detection, color separation) yields finer-grained representations \cite{Sia_Eye_bro}, while dual-stream attention frameworks enhance coupling between local and global cues \cite{TransEye,Tan2023}. Nevertheless, simple feature concatenation and some multi-resolution aggregation schemes remain limited in exploiting multi-scale spatial semantics \cite{CHEN2024106045,10772436}. Overall, prior work advances paired-modal modeling but still suffers incomplete feature fusion and inadequate cross-modal integration.

Motivated by these gaps, this study targets the representational limits of existing Siamese approaches by integrating morphological–semantic spatial feature learning to improve multi-scale fusion and cross-channel interaction.

\subsection{Classification of Binocular Fundus Images Combined with Ophthalmic Robotics}

Fundus image recognition is integral to ophthalmic robotics and telemedicine. Recent reviews summarize AI’s role in surgical scene identification and real-time feedback for improved surgical accuracy and remote operation feasibility \cite{Madanan2025}. Deep learning has also been applied to robotic ocular imaging: Khan et al. (2021) demonstrated the potential of AI-integrated, self‑calibrated OCT systems for automated screening, while noting elevated autonomy requirements for clinical deployment \cite{Khan_Kallogjeri_Piccirillo_2021}. Robotic precision advances have enabled reliable interventions such as an automated device for robot‑assisted retinal vein injection \cite{10.3389/frobt.2022.913930}. In telemedicine, robotic process automation and machine‑learning tools have been shown to save time, improve satisfaction, and reduce clinician workload in remote ophthalmic screening \cite{THAINIMIT2022101001}, but greater intelligence and usability are needed to reduce dependence on expert physicians.

Work on diagnostic robots demonstrates growing capability for non‑specialists: reviews and systems show that AI‑assisted robots can support fundus abnormality detection \cite{10.1167/tvst.14.1.14}, and platforms that automatically acquire ocular-surface and fundus images reduce the need for on-site ophthalmologists, facilitating remote care \cite{Lie077859}. Because diagnostic performance depends on image quality, substantial efforts have improved image capture and robot adaptability—for example, methods to reduce motion artifacts \cite{10585590}, techniques to relax constraints on patient posture and head orientation \cite{10130250}, and mobile‑robot functions such as automatic obstacle avoidance and dynamic tracking to enhance scene adaptability and lower operator demands \cite{11025963}.

\section{The Propsoed Method}
\begin{figure}
    \centering
    \includegraphics[width=\columnwidth]{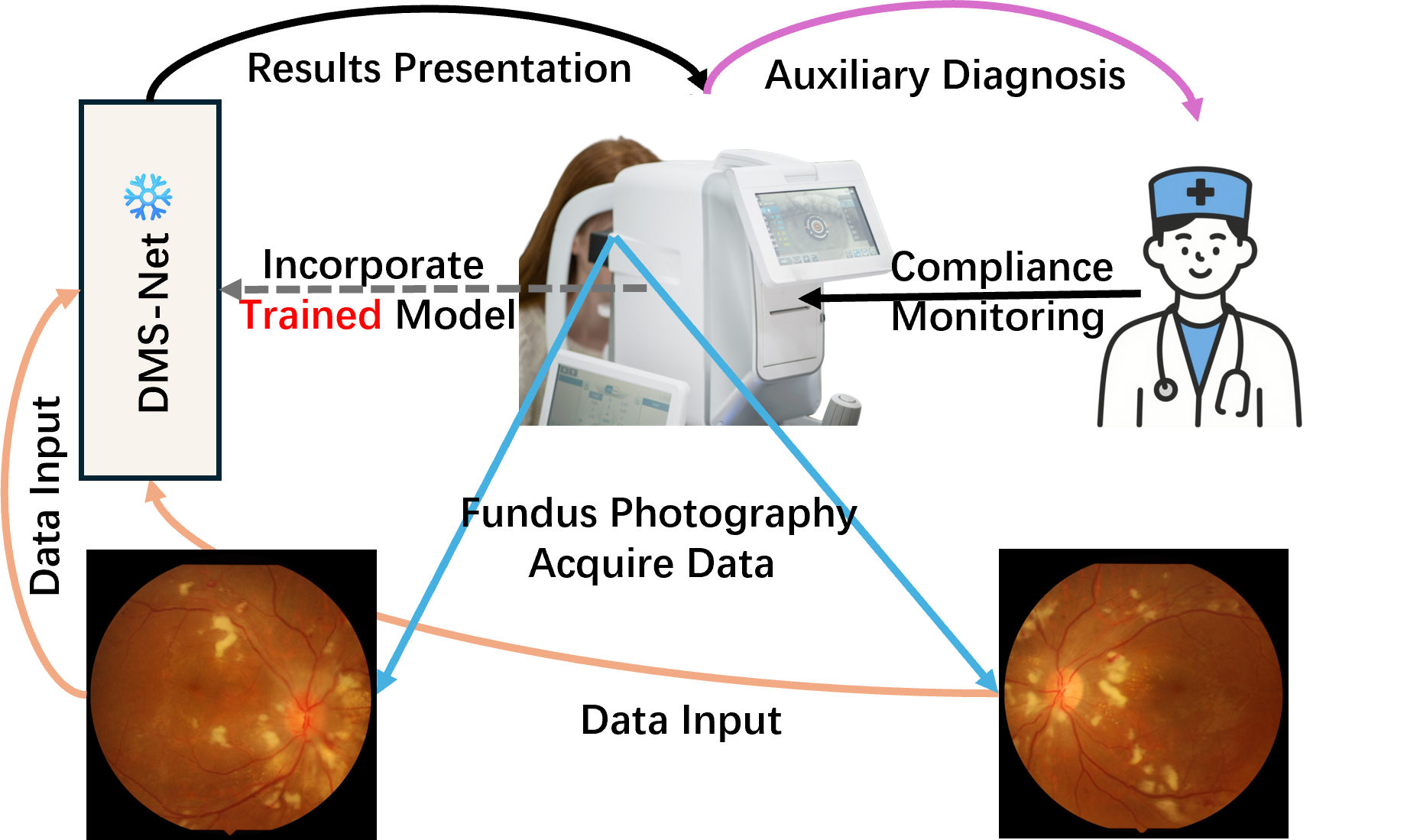}
    \caption{Workflow Diagram of a Fundus Imaging Disease Diagnosis Robot with Integrated DMS-Net. Clinicians continuously monitor the use of the fundus imaging diagnosis robot. The device scans the patient's retina to acquire fundus images, which are then processed by its integrated, trained DMS-Net model. Disease classification results are displayed on-screen to assist in physician diagnosis.}
    \label{fig:0}
    \vspace{-3mm}  
\end{figure}

\begin{figure*}[h!]
    \centering
    \includegraphics[width=1\textwidth]{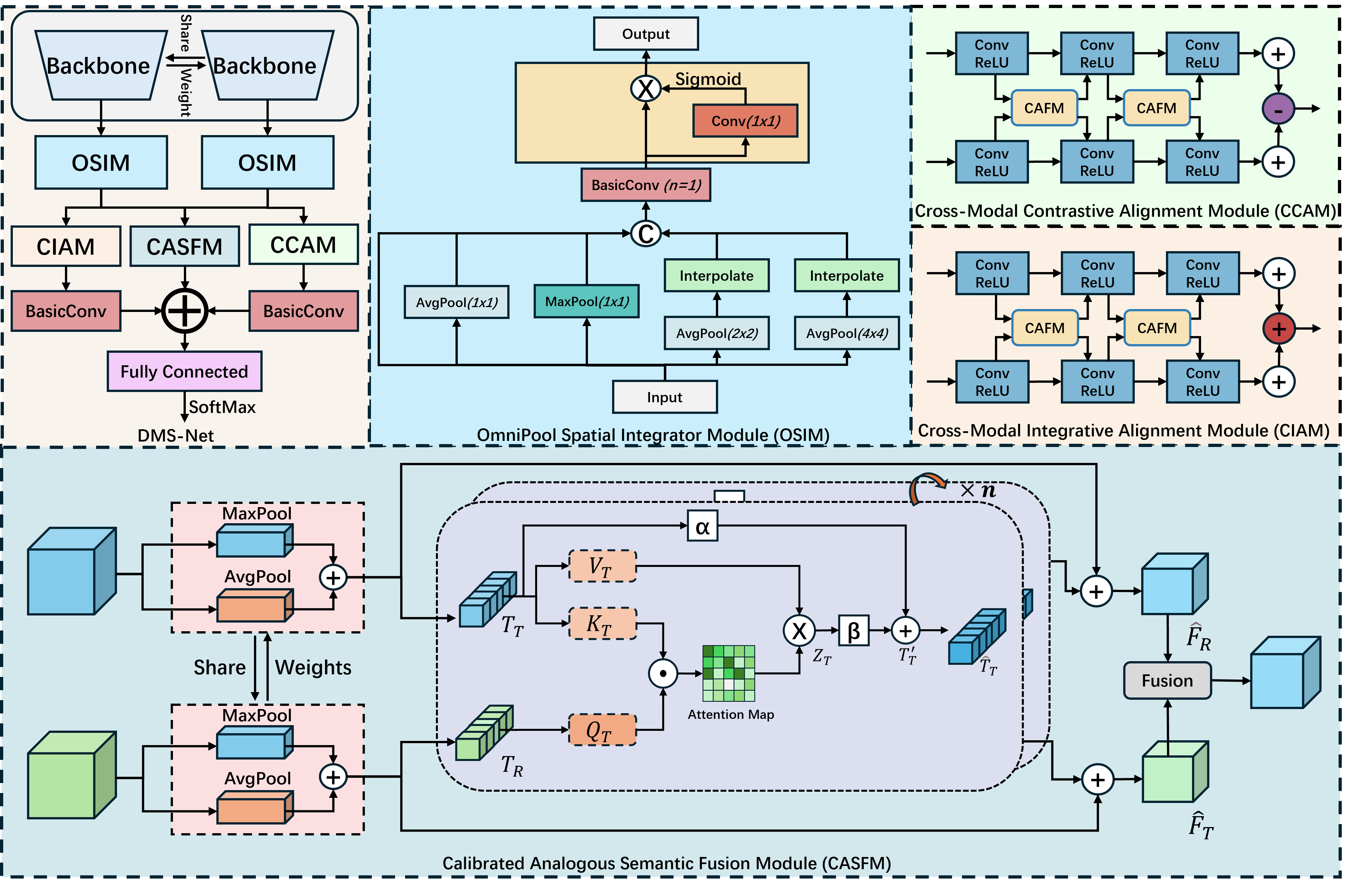}
    \caption{The architecture of DMS-Net.}
    \label{fig:DMS-Net}
\end{figure*}

We propose DMS-Net, a Siamese network-based framework for binocular retinal image (binocular fundus image) classification.  As shown in Figure \ref{fig:0}, our fundus disease diagnosis robot incorporating DMS-Net demonstrates its workflow. As illustrated in Figure \ref{fig:DMS-Net}, the model employs dual weight-sharing ResNet-152 branches to concurrently extract deep semantic features from dual-channel (left-right) ocular images. To effectively address the challenges of indistinct lesion boundaries and morphological heterogeneity, our method innovatively incorporates the OmniPool Spatial Integrator Module (OSIM). OSIM integrates multi-scale adaptive average pooling (Adaptive Average Pooling), global max pooling (Global Max Pooling), and a spatial attention mechanism (Spatial Attention Mechanism).

Furthermore, we designed the Calibrated Analogous Semantic Fusion Module (CASFM), which leverages spatial attention, bidirectional cross-attention mechanisms, and parametrically adaptive residual connections to facilitate cross-modal (inter-channel) spatial-semantic interaction.

To comprehensively identify and effectively leverage the differential semantic information of lesion features present in the left and right retinal images (fundus images), we introduce the Cross-modal Contrastive Alignment Module (CCAM). Concurrently, considering the existence of correlative information for symmetrical pathologies across different lesions in both eyes, we designed the Cross-modal Integrative Alignment Module (CIAM) to enhance the deep aggregation of lesion-correlated information extracted from the dual channels, thereby generating more discriminative representations.

\subsection{Siamese Network Feature Extraction}

In this study, we constructed a binocular retinal image (fundus image) feature extraction backbone network based on a pre-trained ResNet-152 architecture. Implementing a weight-sharing strategy to achieve synchronized parameter updates, the network enables parallel extraction of deep semantic features from paired left-right retinal (fundus) images. This design effectively captures morphological feature correlations in lesion regions across both fundus images, alongside their high-level semantic representations.

\subsection{OmniPool Spatial Integrator Module (OSIM)}

To address the technical challenges of boundary ambiguity, scattered lesion distribution, and significant morphological variations in retinal lesion images, this study proposes an innovative OmniPool Spatial Integrator Module (OSIM). The module architecture diagram is shown in Figure \ref{fig:DMS-Net}. The module builds upon high-level semantic features extracted by the ResNet152 network and employs a multi-scale pooling-based feature enhancement strategy: Adaptive Average Pooling with a scale parameter of 2 captures medium-grained semantic representations, while adaptive pooling operations with a scale parameter of 4 extract coarse-grained semantic features:
\begin{align}
    \mathrm{SPP}_{2\times2}(x) = \mathrm{interp}\left( \mathrm{avgpool}_{(2,2)}(x), \mathrm{size}=H \times W \right)
\end{align}%

\begin{align}
    \mathrm{SPP}_{4\times4}(x) = \mathrm{interp}\left( \mathrm{avgpool}_{(4,4)}(x), \mathrm{size}=H \times W \right)
\end{align}%
In the formula, $x \in \mathbb{R}^{H \times W \times C}$ denotes the high-level semantic features extracted by the ResNet152 network. Here, $\mathrm{interp}$ represents bilinear interpolation $\mathrm{mode} = \mathrm{bilinear}$ to restore the feature map to its original dimensions $H \times W $. This is followed by the integration of Global Max Pooling to enhance feature responses in salient lesion regions:
\begin{align}
    \text{GMP}(x) = \mathrm{maxpool}_{(1,1)}(x) \odot \mathbf{1}
\end{align}%
To capture global contextual semantic information, we also introduce Global Average Pooling: 
\begin{align}
    \text{GAP}(x) = \mathrm{avgpool}_{(1,1)}(x) \odot \mathbf{1}
\end{align}%
In the formula, $\odot$ denotes element-wise multiplication, where the tensor 1 (all-ones) is broadcast to match the original dimensions $H \times W $. To alleviate the potential loss of subtle lesion features caused by pooling operations, a supplementary mechanism is specifically designed to retain the original high-level semantic features.
\begin{equation} \label{eq:pool}
    \begin{aligned}
        F_{input} = \text{Concat}\bigl( &x, \text{GMP}(x), \text{GAP}(x), \\
        &\mathrm{SPP}_{(2\times2)}(x), \mathrm{SPP}_{(4\times4)}(x), \text{dim}=1 \bigr)
    \end{aligned}
\end{equation}
Specifically, the$\text{Concat}$ operation is employed to fuse semantic features of five different scales along the channel dimension, resulting in a combined feature map $X \in \mathbb{R}^{H \times W \times 5C}$. This fused representation is then fed into a channel compression mechanism for adaptive channel-wise fusion.
\begin{align}
    F_{ccp}=\mathrm{ReLU} ( \mathrm{BatchNorm} ( \mathrm{Conv2d_{(1\times1)}}\left(F_{input}\right)))
\end{align}%
The five types of multi-scale features (including the original high-level semantic features, two-scale pooling features, and two global pooling features) are adaptively fused through 2D convolution operations to obtain $F_{ccp} \in \mathbb{R}^{H \times W \times (C/2)}$. Finally, a spatial attention mechanism is applied to strengthen the feature responses in lesion regions.
\begin{align}
    F_{out} = F_{ccp} \odot \sigma\left( W_s \ast F_{ccp} + b_s \right)
\end{align}%
In the formula, $\sigma$denotes the $\mathrm{Sigmoid}$ activation function, and $F_{out} \in \mathbb{R}^{H \times W \times (C/2)}$ represents the output of the multi-scale feature extraction module integrated with a spatial attention mechanism. This integration aims to achieve focused enhancement of critical pathological features.

\subsection{Calibrated Analogous Semantic Fusion Module (CASFM)}

To enhance the performance of Siamese networks in spatial-semantic representation of lesion localization and tissue structure correlation, this study designs a Calibrated Analogous Semantic Fusion Module (CASFM). The architecture diagram of the CASFM is as shown in Figure \ref{fig:DMS-Net}. This module constructs a modality-agnostic semantic representation system for lesions through multi-scale cross-modal feature spatial-semantic extraction and cross-modal fusion. Specifically, the CASFM module adopts a dual-path parallel processing framework:

First, max-pooling and average-pooling operations are applied to the input binocular features to achieve spatial dimension compression. 
\begin{equation} \label{eq:pool}
    \begin{split}
        F_{left/right}^{avg} &= \frac{1}{k^2}\mathrm{avgpool}(\mathrm{Conv2d}_{(1\times1)}(F_{left/right})) \\
        F_{left/right}^{max} &= \frac{1}{k^2}\mathrm{maxpool}(\mathrm{Conv2d}_{(1\times1)}(F_{left/right}))
    \end{split}
\end{equation}

In the formula, $k$ denotes the size of the pooling kernel. We set $k=2$, and the input features $F_{right/left}\in \mathbb{R}^{H \times W \times (C/2)}$ correspond to the output $F_{out}$ from the multi-scale feature extraction module. Multi-granularity feature fusion is implemented via learnable modality-adaptive weighting factors $\lambda_{left/right} \in [0,1]$, effectively preserving global contextual information and local detail features of lesions:

\begin{equation} \label{eq:pool}
    \begin{split}
        F_{left} &= \lambda_{left} \cdot F_{left}^{max} + (1 - \lambda_{left}) \cdot F_{left}^{avg} \\
        F_{right} &= \lambda_{right} \cdot F_{right}^{max} + (1 - \lambda_{right}) \cdot F_{right}^{avg}
    \end{split}
\end{equation}

To mitigate interference from spatial positional deviations of binocular fundus high-level semantic features on lesion semantic alignment, modality-specific spatial embedding vectors are introduced for positional encoding compensation:
\begin{equation} \label{eq:pool}
    \begin{split}
        F_{left/right}^{\prime} &= \text{Flatten}(\mathrm{Conv2d}_{(1\times1)}(F_{left/right}) \\
        &\quad + \text{Interp}(P_0, (H, W)) \cdot W_p 
    \end{split}
\end{equation}

In the formula, $P_0 \in \mathbb{R}^{C_{\text{emb}} \times 1 \times 1}$denotes the initial positional embedding parameters. A bidirectional multi-head attention mechanism establishes cross-modal feature interaction channels, achieving collaborative enhancement of lesion semantics across different subspaces:

\begin{equation} \label{eq:pool}
    \begin{split}
        Z_{right} &= \mathrm{softmax}(\frac{Q_{left}K_{right}^{T}}{\sqrt{D_{K}}})\cdot V_{right} \\
        Z_{left} &= \mathrm{softmax}(\frac{Q_{right}K_{left}^{T}}{\sqrt{D_{K}}})\cdot V_{left}
    \end{split}
\end{equation}

A parameter-adaptive residual connection strategy fuses enhanced features with original features:

\begin{equation} \label{eq:pool}
    \begin{split}
        T_{left}^{\prime} &= \alpha_{left}\cdot Z_{left}\cdot W^O+\beta_{left}\cdot T_{left} \\
        T_{right}^{\prime} &= \alpha_{right}\cdot Z_{right}\cdot W^O+\beta_{right}\cdot T_{right}
    \end{split}
\end{equation}

In the formula, $\alpha_{left/right} \in [0,1]$ and $\beta_{left/right} \in [0,1]$denote adaptive fusion parameters. 

Finally, the modality-disentangled dual-branch spatial-semantic features are adaptively fused to reconstruct modality-agnostic spatial-semantic feature information:

\begin{align}
    F_{fuse}=\mathrm{Conv2d}_{(1\times1)}(\mathrm{Concat}(\hat{T}_{left}, \hat{T}_{right}))
\end{align}%

\subsection{Dual-Synergy Cross-Modal Fusion}

To better align and fuse the semantic relationships between features of binocular retinal (fundus) images, we designed the Cross-Modal Contrastive Alignment Module (CCAM) and the Cross-Modal Integrative Alignment Module (CIAM) to operate in parallel. As shown in Figure \ref{fig:DMS-Net}, CCAM captures differential pathological features between left-right ocular images, while CIAM aggregates and enhances interocular lesion-correlated semantic features.

\textbf{Cross-Modal Contrastive Alignment Module (CCAM)} In conditions like diabetic retinopathy, binocular (bilateral) ocular pathology often manifests asymmetrically (e.g., significant hemorrhage in the left eye versus mild pathology in the right eye). The CCAM leverages the Cross-Attention Fusion Module (CAFM) to compute the distributional discrepancy between cross-eye features. It then guides the dual-path densely connected convolutional network to progressively achieve pathological-asymmetry-aligned semantic representations of binocular lesions. This mechanism directs the model to concentrate on asymmetry characteristics among lesions, such as disparities in hemorrhage extent and exudate morphology.

\textbf{Cross-Modal Integrative Alignment Module (CIAM)} In diseases such as glaucoma and hypertensive retinopathy, there exist symmetrically co-evolving pathological patterns (e.g., synchronous enlargement of the cup-to-disc ratio, symmetric increase in vascular stenosis severity). The CIAM leverages the Cross-Attention Fusion Module (CAFM) to mine and quantify interocular symmetrical pathological correlations, exemplified by computing spatial similarity of feature maps across key anatomical regions (e.g., optic disc localization), and achieves lesion region alignment (like optic disc co-registration). Subsequently, the module guides the dual-path densely connected convolutional network towards synergistic enhancement of binocular pathological semantic features, while employing residual connections to preserve the original feature information and mitigate critical information loss.

\section{Dataset}
ODIR-5K \cite{dataset} is a structured medical imaging dataset dedicated to ophthalmic disease intelligent recognition. Jointly released by Peking University and institutions including Shanghai Medical Technology in 2019, it aims to advance the application of artificial intelligence in fundus disease classification. The dataset contains anonymized clinical data from 5,000 patients, with each patient record encompassing gender, age, color fundus photographs of both eyes, and physicians' diagnostic keywords.
\subsection{Data process}
\begin{figure}[h!]
    \centering
    \includegraphics[width=\columnwidth]{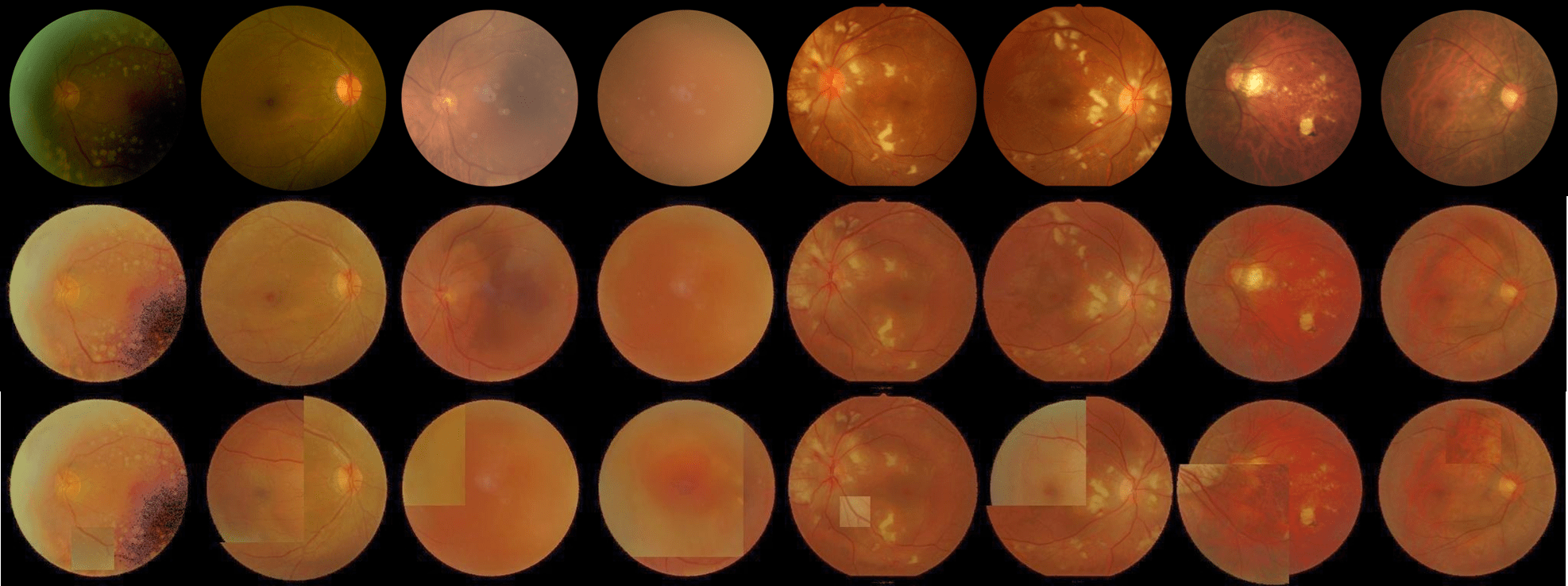}
    \caption{The first row displays four pairs of original fundus imaging data, sequentially presenting cases of diabetic retinopathy, cataract, hypertensive retinopathy, and high myopia. The second row shows enhanced images after non-uniform illumination correction processing. The third row demonstrates newly synthesized data generated by applying CutMix augmentation on the illumination-corrected images.}
    \label{fig:1}
    \vspace{-3mm}  
\end{figure}

\begin{figure}[h!]
    \centering
    \includegraphics[width=\columnwidth]{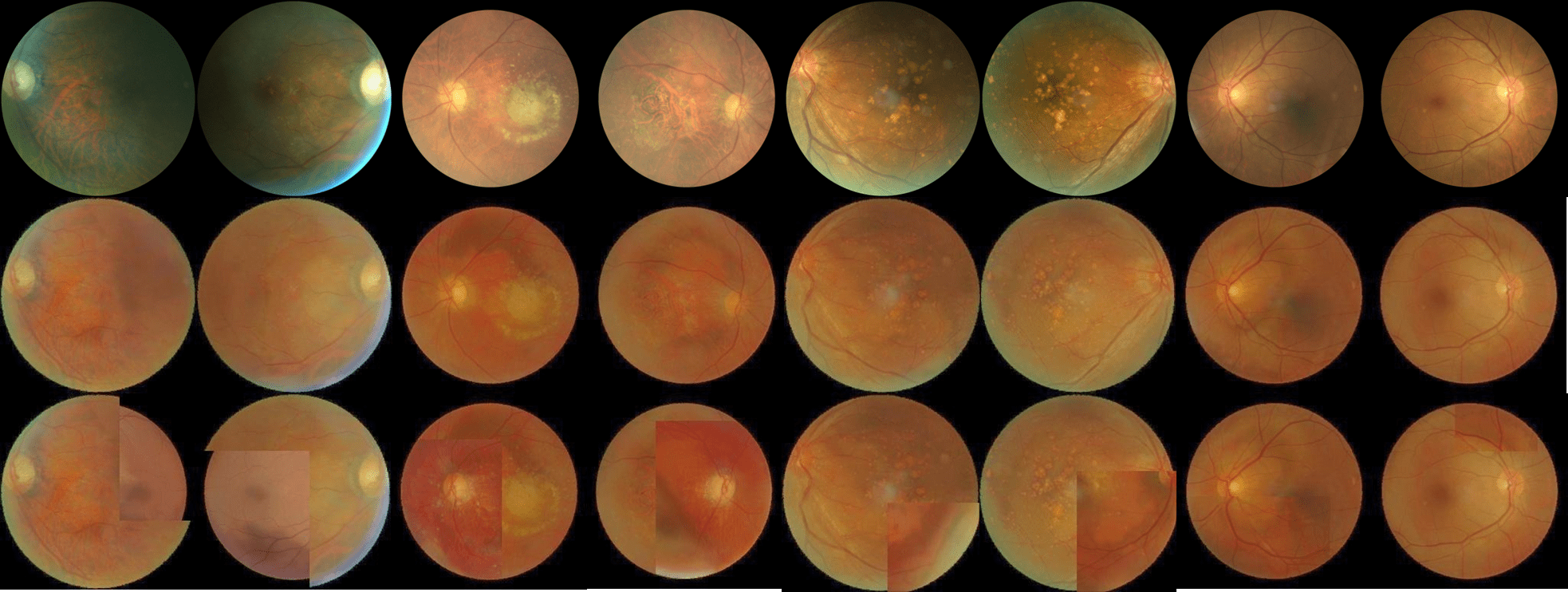}
    \caption{The first row displays four pairs of original fundus imaging data, from left to right, respectively representing other ocular pathologies, glaucoma, macular degeneration, and normal retinal structures. The second row demonstrates enhanced images after non-uniform illumination correction processing. The third row exhibits newly synthesized data generated by implementing CutMix augmentation on the illumination-corrected images.}
    \label{fig:2}
    \vspace{-3mm}  
\end{figure}

We employed non-uniform illumination correction preprocessing for image optimization, combined with the CutMix data augmentation algorithm to integrate local regions of homogeneous samples while preserving original class labels. This strategy effectively guides the model to focus on learning more discriminative local features, achieving diversified dataset expansion while maintaining lesion region consistency, thereby enhancing model generalization capability. Considering the class imbalance in the original dataset objectively reflects epidemiological characteristics of real-world disease distribution, a class distribution preservation strategy was implemented during data preprocessing to moderately maintain the inherent class imbalance. This approach ensures the model effectively learns clinically representative data distribution patterns. Figures \ref{fig:1} and \ref{fig:2} present our raw data and the step-by-step data processing visualization results.

\begin{figure}[h!]
    \centering
    \includegraphics[width=\columnwidth]{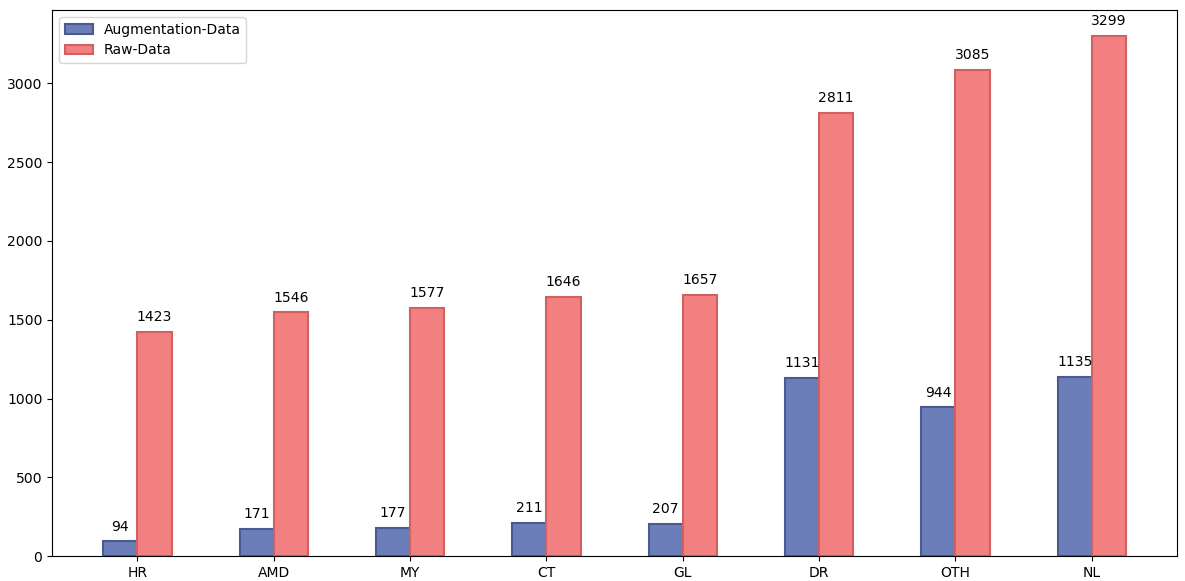}
    \caption{Comparison of raw data versus augmented data quantities. 
             Note the significant increase in sample size after augmentation.}
    \label{fig:3}
    \vspace{-3mm}  
\end{figure}
The figure \ref{fig:3} illustrates the distribution of fundus image data across various ophthalmic disease categories (including normal samples) in the original dataset and their corresponding quantities after data processing. The abbreviations on the bar chart's horizontal axis correspond to the following ophthalmic conditions: N (NL) denotes Normal, D (DR) represents Diabetic Retinopathy, G (GL) indicates Glaucoma, C (CT) corresponds to Cataract, A (AMD) signifies Age-related Macular Degeneration, H (HR) refers to Hypertensive Retinopathy, M (MY) stands for Myopia, and O (OTH) designates Other Disease. This standardized labeling system uses initial letter abbreviations to maintain clarity in visualizing epidemiological data distributions across disease categories.

\section{Ablation Study}
We utilize Accuracy (Acc) to quantify the overall correctness of model predictions, Precision to evaluate the reliability of positive-class predictions, and Recall to assess the model's ability to capture true positive samples. The Kappa Coefficient measures the statistical divergence between classification results and random predictions, while the F1 Score balances Precision and Recall through a harmonic mean. AUC (Area Under the ROC Curve) further characterizes the model's capability to distinguish between positive and negative classes. By integrating these multidimensional metrics, we comprehensively evaluate the model's performance.

We conducted experiments to explore the following two aspects: (1) the model performance using different backbones as feature extractors, and the experimental results are shown in Table \ref{tab:ab1}; (2) the effectiveness of each component of the model, and the experimental results are shown in Table \ref{tab:ab2}.

\subsection{Model Performance with Different Backbones}
The ablation study comparing the impact of different backbone networks on model performance reveals that ResNet-152 consistently achieves the highest performance across all metrics(as shown in Table \ref{tab:ab1}), including accuracy, precision, recall, Kappa coefficient, F1-score, and area under the curve (AUC), with an accuracy of 0.829 and an AUC of 0.973, indicating its superiority in feature extraction and classification stability; in contrast, the Vision Transformer (ViT) demonstrates relatively weaker performance with lower metrics (e.g., accuracy at 0.804), highlighting its limitations for the task, while the ResNet series exhibits progressive improvement in performance with increasing depth (e.g., from ResNet-50 to ResNet-152), underscoring the positive effect of model complexity on precision.

\begin{table}[!htbp]
    \centering
    \caption{Model Performance with Different Backbones}
    \vspace{-3mm}  
    \scriptsize  
    \begin{tabular}{c|c|c|c|c|c|c}
        \toprule
        Backbone & Acc & Precision & Recall & Kappa & F1 & AUC  \\
        \midrule
        ViT & 0.804 & 0.826 & 0.826 & 0.797 & 0.825 & 0.949 \\
        ResNeXt & 0.792 & 0.827 & 0.806 & 0.784 & 0.812 & 0.962 \\
        ResNet-50 & 0.817 & 0.854 & \underline{0.827} & 0.812 & 0.837 & 0.962 \\
        ResNet-101 & \underline{0.819} & \underline{0.860} & 0.825 & \underline{0.814} & \underline{0.839} & \underline{0.965} \\
        ResNet-152 & \textbf{0.829} & \textbf{0.869} & \textbf{0.845} & \textbf{0.832} & \textbf{0.856} & \textbf{0.973}\\
        \bottomrule
    \end{tabular}
    \label{tab:ab1}
    \vspace{2mm}  
\end{table}

\subsection{Investigation into the Effectiveness of Model Components}
As delineated in Table \ref{tab:ab2}, quantitative ablation studies rigorously validate the contribution of each module. Removing the CASFM causes the most severe degradation: Kappa coefficient declines by 2.5\% (0.832 to 0.807), accuracy drops 1.7\% (0.829 to 0.812), with an equivalent 1.7\% recall reduction (0.845 to 0.828). The dual-synergy modules also demonstrate critical roles: eliminating the CCAM reduces accuracy by 1.0\% (0.829 to 0.819), whereas removing the CIAM leads to a 1.3\% accuracy decrease (0.829 to 0.816). The integrity of the OSIM is vital for model stability, as ablation of its left sub-module decreases Kappa by 1.3\% (0.832 to 0.819), and right sub-module removal results in 2.0\% Kappa reduction (0.832 to 0.812). The full model (ALL) achieves optimal performance across all metrics, with its 0.973 AUC and 0.856 F1-score substantially surpassing any ablated configuration.

\begin{table}[!htbp]
    \centering
    \caption{Investigation into the Effectiveness of Model Components}
    \vspace{-3mm}  
    \scriptsize  
    \begin{tabular}{c|c|c|c|c|c|c}
        \hline
        Configuration & Acc & Precision & Recall & Kappa & F1 & AUC  \\
        \hline
        w/o CAFM &  0.819 & 0.854 & 0.829 & 0.815 & 0.840 & \underline{0.967} \\
        w/o CIAM & 0.816 & 0.856 & 0.829 & 0.815 & 0.841 & 0.962 \\
        w/o CCAM &  0.819 & \underline{0.859} & 0.836 & \underline{0.822} & \underline{0.848} & 0.963 \\
        w/o CASFM & 0.812 & 0.844 & 0.828 & 0.807 & 0.834 & 0.960 \\
        w/o OSIM(Left)    & 0.816 & 0.853 & \underline{0.837} & 0.819 & 0.845 & 0.961\\
        w/o OSIM(Right)    & 0.818 & 0.850 & 0.830 & 0.812 & 0.838 & 0.963\\
        w/o OSIM(ALL)    & \underline{0.822} & 0.857 & 0.835 & 0.817 & 0.843 & 0.965\\
        ALL & \textbf{0.829} & \textbf{0.869} & \textbf{0.845} & \textbf{0.832} & \textbf{0.856} & \textbf{0.973}\\
        \hline
    \end{tabular}
    \label{tab:ab2}
\end{table}

\subsection{Model Performance with Different Feature Extraction Module}

We investigated the impact of different plug-and-play multi-scale modules on model performance, as shown in Table \ref{tab:ab0}, where our OSIM module was replaced with ASPP \cite{ASPP}, SPPF \cite{yolov5}, SPP \cite{SPP}, RFB \cite{RFB}, and SimSPPF \cite{yolov6} variants. Ablation experiments confirmed that the proposed OSIM module achieves superior performance compared to these alternatives.

\begin{table}[!htbp]
    \centering
    \caption{Model Performance with Different Feature Extraction Module.}
    \vspace{-3mm}  
    \scriptsize  
    \begin{tabular}{c|c|c|c|c|c|c}
        \hline
        Backbone & Acc & Precision & Recall & Kappa & F1 & AUC  \\
        \hline
        ASPP & 0.776 & 0.816 & 0.793 & 0.769 & 0.800 & 0.958 \\
        SPPF &  0.774 & 0.828 & 0.799 & 0.778 & 0.807 & 0.964 \\
        SPP & 0.805 & 0.836 & 0.821 & 0.800 & 0.827 & 0.965 \\
        simSPPF & 0.792 & 0.827 & 0.807 & 0.784 & 0.812 & 0.961  \\
        RFB  & \underline{0.824} & \underline{0.867} & \underline{0.832} & \underline{0.828} & \underline{0.852} & \underline{0.971}\\
        OSIM (Ours) & \textbf{0.829} & \textbf{0.869} & \textbf{0.845} & \textbf{0.832} & \textbf{0.856} & \textbf{0.973}\\
        \hline
    \end{tabular}
    \label{tab:ab0}
\end{table}

\section{Conclusion}
This study proposes DMS-Net for binocular retinal image classification – a dual-modal multi-scale siamese network framework. By integrating a weight-sharing Siamese ResNet-152 backbone, the multi-scale context aggregation module (OSIM), and the Calibrated Analogous Semantic Fusion Module (CASFM), the framework effectively addresses core challenges in binocular fundus image analysis: indistinct lesion boundaries, diffuse pathological distributions, and cross-modal (dual-channel) information interaction. OSIM achieves robust extraction of multi-resolution pathological features through its hierarchical multi-scale pooling structure (improving Kappa coefficient by 0.8\% over the best baseline). CASFM leverages spatial attention and feature recalibration mechanisms to capture and enhance discriminative, modality-agnostic pathological semantic representations. Crucially, the innovatively designed Cross-modal Contrastive Alignment Module (CCAM) and Cross-modal Integrative Alignment Module (CIAM) achieve, for the first time, a synergistic modeling mechanism for asymmetric lesion manifestations (e.g., unilateral hemorrhage) and symmetrically evolving pathological patterns (e.g., bilateral synchronous changes in cup-to-disc ratio). This study contributes to enhancing the diagnostic precision and reliability of existing ophthalmic robots when processing complex fundus images characterized by asymmetric lesions, indistinct boundaries, and other complexities. Evaluated on the ODIR-5K dataset, DMS-Net sets a new state-of-the-art performance with an accuracy of 82.9\%, a recall of 84.5\%, and a Cohen's Kappa coefficient of 83.2\%.


\bibliography{ref}

\end{document}